\pdfoutput=1

\documentclass[11pt]{article}

\usepackage[]{EACL2023}

\usepackage{times}
\usepackage{latexsym}

\usepackage[T1]{fontenc}

\usepackage[utf8]{inputenc}

\usepackage{microtype}

\usepackage{inconsolata}

\usepackage{booktabs}
\usepackage{lipsum}
\usepackage{graphicx}
\usepackage{amssymb}

\usepackage{enumitem}
\usepackage{supertabular}

\newcommand{\token}[1]{\textcolor{darkblue}{\textbf{\small{#1}}}}
\definecolor{celestialblue}{rgb}{0.29, 0.59, 0.82}
\definecolor{cerulean}{rgb}{0.0, 0.28, 0.67}

\title{Behavior Cloned Transformers are Neurosymbolic Reasoners} 

\author{Ruoyao Wang$^{\ddagger}$, Peter Jansen$^{\ddagger}$, Marc-Alexandre Côté$^{\clubsuit}$,
Prithviraj Ammanabrolu$^{\diamondsuit}$ \\
$^{\ddagger}$University of Arizona, Tucson, AZ  ~~~~~ $^{\clubsuit}$Microsoft Research Montréal \\
$^{\diamondsuit}$Allen Institute for AI, Seattle, WA \\
\texttt{\{ruoyaowang,pajansen\}@arizona.edu}\\ \texttt{macote@microsoft.com}, \texttt{raja@allenai.org}
}

\begin{document}
\maketitle
\begin{abstract}
In this work, we explore techniques for augmenting interactive AI agents with information from symbolic modules, much like humans use tools like calculators and GPS systems to assist with arithmetic and navigation.
We test our agent's abilities in text games---challenging benchmarks for evaluating the multi-step reasoning abilities of game agents in grounded, language-based environments.  
Our experimental study indicates that injecting the actions from these symbolic modules into the action space of a behavior cloned transformer agent increases performance on four text game benchmarks that test arithmetic, navigation, sorting, and common sense reasoning by an average of 22\%, allowing an agent to reach the highest possible performance on unseen games. This action injection technique is easily extended to new agents, environments, and symbolic modules.\footnote{We release our system as open source, available at \url{http://github.com/cognitiveailab/neurosymbolic/}}

\end{abstract}

\section{Introduction}
Interactive fiction games (or \textit{text games)} evaluate AI agents abilities to perform complex multi-step reasoning tasks in interactive environments that are rendered exclusively using textual descriptions. Agents typically find these games challenging due to the complexities of the tasks combined with the reasoning limitations of contemporary models.  Overall performance is generally low, with agents currently solving only 30\% of classic interactive fiction games such as Zork \cite{ammanabrolu2020graph,yao-etal-2021-reading,atzeni2022casebased}.  
Similarly, reframing benchmarks such as question answering into text games where agents must interactively reason with their environment and make their reasoning steps explicit causes performance to substantially decrease \cite{wang2022scienceworld}, highlighting both the capacity of this methodology to evaluate multi-step reasoning, and the limitations of current language models.

\begin{figure}[t!]
\centering
\includegraphics[scale=0.85]{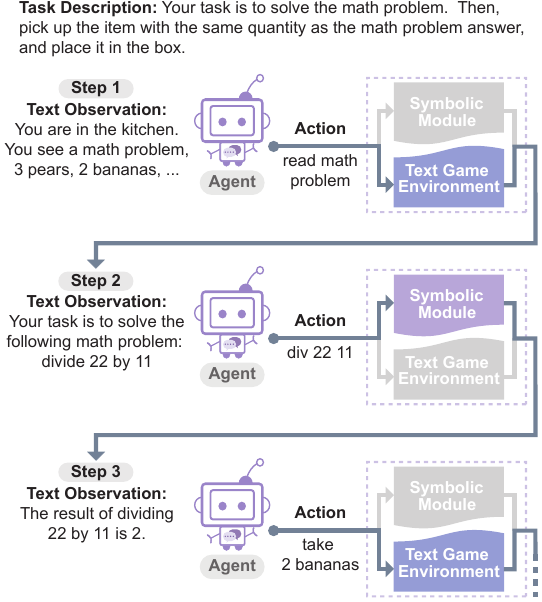}
\caption{An overview of our approach on an example game evaluating arithmetic ability.  At each step, the agent receives an observation from the environment, then takes an action.  By providing actions that interface to symbolic modules (such as a calculator), the agent is able to use external knowledge to help solve the task. }
\label{fig:overview}
\vspace{-4mm}
\end{figure}

While large language models are capable of a variety of common sense reasoning abilities \cite{liu-etal-2022-generated, ji-etal-2020-language}, contemporary agents typically struggle on tasks such as navigation, arithmetic, knowledge base lookup, and other tasks that humans typically make use of external tools (such as GPS systems, calculators, and books) to solve.  This is at times frustrating, because 
the tasks they perform poorly on can sometimes be solved in a few dozen lines of code.  In this work, we show that combining both approaches is possible for text game agents, with our approach shown in Figure~\ref{fig:overview}.  We develop symbolic modules for arithmetic, navigation, sorting, and knowledge base lookup in \textsc{Python}, paired with new benchmark games for testing these capacities in interactive text game environments.  We empirically demonstrate that injecting actions from those modules into the action space of each game can allow transformer-based agents to make use of that information, and achieve near-ceiling performance on unseen benchmark games that they otherwise find challenging.  

\section{Related Work}

Neurosymbolic reasoning offers the promise of combining the inference capabilities of symbolic programs with the robustness of large neural networks. 
In the context of text games, Kimura et al.~\shortcite{kimura-etal-2021-loa} develop methods to decompose text games into a set of logical rules, then combine these rules with deep reinforcement learning \cite{kimura-etal-2021-neuro} or integer linear programming \cite{basu2021hybrid} to substantially increase agent performance 
 while providing a more interpretable framework for understanding why agents choose specific actions \cite{chaudhury-etal-2021-neuro}.  
 More generally, neurosymbolic reasoning has been applied to a variety of multi-step inference problems, such as multi-hop question answering \cite{weber-etal-2019-nlprolog}, language grounding \cite{zellers-etal-2021-piglet}, and semantic analysis \cite{cambria-etal-2022-senticnet}.

Because text games require interactive multi-step reasoning, agents have most commonly been modelled using reinforcement learning \cite[e.g.][]{he-etal-2016-deep-reinforcement, zahavy2018learn, yao-etal-2020-keep}, though overall performance on most environments remains low \cite[see][for reviews]{jansen-2022-systematic,Osborne2021ASO}.
Recently, alternative approaches modeling reinforcement learning as a sequence-to-sequence problem using imitation learning have emerged, centrally using behavior cloning \cite{ijcai2018p687}, decision transformers \cite{chen2021decision}, and trajectory transformers \cite{janner2021sequence}.  These approaches model interactive multi-step reasoning problems as a Markov decision process, where an agent's observation and action history up to some depth are provided as input, and the transformer must predict the next action for the agent to take.  Behavior cloning and decision transformers have recently been applied to text games with limited success \cite{wang2022scienceworld}.  Here, we show that the performance of a behavior cloned transformer can substantially increase when augmented with neurosymbolic reasoning.




\section{Approach}

Figure~\ref{fig:overview} illustrates the workflow of our approach. At each time step $t$, based on the observation $o_t$, the symbolic module will generate a set of valid actions $A^{m}_t$, and the text game environment will have a distinct set of valid actions $A^{e}_t$. Let the valid action set at step $t$ be $A_t = A^{m}_t \cup A^{e}_t$. Given $o_t$ and $A_t$, the agent needs to choose an action $a_t \in A_t$ to take. Note that in principle, any agent could be adapted to use this approach, since we simply inject actions from the symbolic modules into the environment action space.  At a given time step, our approach checks if $a_t$ is a valid symbolic action. If $a_t \in A^{m}_t$ (e.g. \textit{div 22 11} in Figure~\ref{fig:overview}), the symbolic module will generate the next observation $o_{t+1}$, otherwise the text game environment will take $a_t$ and generate $o_{t+1}$ (e.g. \textit{take 2 bananas} in Figure~\ref{fig:overview}).

\section{Environments and Symbolic Modules}
We evaluate our approach to neurosymbolic reasoning using four text game benchmark environments centered around pick-and-place tasks, including one existing benchmark and three new developed for this work.  Each environment supports parametric variation to generate many different games.  These environments are outlined below, with additional details and example playthroughs found in \textsc{Appendix~\ref{sec:env}}.  All environments were implemented using the \textsc{TextWorldExpress} game engine \cite{jansen2022textworldexpress}. 

{\flushleft\textbf{Text World Common Sense (TWC):}} A benchmark common sense reasoning task \cite{murugesan2021textbased} where agents must collect objects from the environment (e.g. \textit{dirty socks}), and place those objects in their canonical common sense locations (e.g. \textit{washing machine}).  The symbolic module for this game allows agents to query a knowledge base of \textit{(subject, relation, object)} triples (e.g. \textit{(cushion, hasCanonicalLocation, sofa)}).

{\flushleft\textbf{MapReader:}} A navigation-themed pick-and-place game similar to Coin Collector \cite{Yuan2018CountingTE}.  An agent starts in a random location (e.g. \textit{the kitchen}), and is provided with a target location (e.g. \textit{the garage}).  The agent must navigate to the target location, pick up a coin, then return to the starting location and place it in a box.  The agent is further provided with a map that can be used for efficient route planning.  The navigation symbolic module paired with this environment scrapes the observation space for location information (e.g. \textit{you are currently in the kitchen}), and both complete (e.g. \textit{the map}) or partial (e.g. \textit{to the north you see the hallway}) spatial connection information. 

{\flushleft\textbf{Arithmetic:}} A math-themed task, where agents must read and solve a math problem in order to know which object from a set of objects to pick-and-place.  An example problem is \textit{``take the bundle of objects that is equal to 3 multiplied by 6, and place them in the answer box''}, where the agent must complete the task by choosing \textit{18 apples}. Distractor objects are populated with quantities that correspond to performing the arithmetic incorrectly (e.g. \textit{3 oranges}, corresponding to subtracting 3 from 6).  We pair the arithmetic game with a calculator module capable of performing addition, subtraction, multiplication, and division. 

{\flushleft\textbf{Sorting:}} A sorting-themed game where the agent begins in a room with three to five objects, and is asked to place them in a box one at a time in order of increasing quantity.  To add complexity, quantities optionally include units (e.g. \textit{5kg of copper, 8mg of steel}) across measures of volume, mass, or length.  The sorting game is paired with a module that scrapes the observation space for mentions of objects that include quantities, and sorts these in ascending or descending order on command.

%
%
\begin{table}[t]
    \centering
    \footnotesize
    \begin{tabular}{l}
    \toprule
    \textbf{Knowledge Base Module}   \\
    \midrule 
    ~~~ \textit{> query cushion}                     \\
    ~~~ cushion located sofa        \\
    ~~~ cushion located armchair    \\
    \midrule
    \textbf{Navigation Module}       \\
    \midrule
    ~~~ You are currently in the kitchen. \\
    ~~~ \textit{> next step to living room} \\
    ~~~ The next location to move to is: hallway. \\
    \midrule
    \textbf{Arithmetic Module}       \\
    \midrule
    ~~~ \textit{> mul 3 6} \\
    ~~~ Multiplying 3 and 6 results in 18. \\
    \midrule
    \textbf{Sorting Module}          \\
    \midrule
    ~~~ \textit{> sort ascending} \\
    ~~~ The objects in ascending order are: 8mg of steel, \\
    ~~~ 2g of iron, 5kg of copper. \\
    \bottomrule
    \end{tabular}
    \caption{Example actions (inputs) and responses from the four symbolic modules investigated in this work.}
    \label{tab:moduleexamples}
\end{table}

\subsection{Symbolic Modules}
Examples of symbolic modules and their responses are provided in Table~\ref{tab:moduleexamples}. The number of valid actions injected by each module varies between 2 from the sorting module (\textit{ascending/descending}) to over 500 from the knowledge base look-up (one for each object and its canonical locations present in the knowledge base). Symbolic modules were implemented in \textsc{Python} as a wrapper around the \textsc{TextWorldExpress} API, allowing modules to monitor observations from the environment, inject actions, and provide responses for any actions they recognized as valid. 

%
%
\begin{table*}[t]
    \centering
    \footnotesize
    \begin{tabular}{lp{5mm}ccccp{5mm}cccc}
    \toprule
                & & \multicolumn{4}{c}{DRRN}    & &   \multicolumn{4}{c}{Behavior Cloned Transformer}                    \\
                 & &  \multicolumn{2}{c}{Baseline}       &   \multicolumn{2}{c}{NeuroSymbolic}            & &      \multicolumn{2}{c}{Baseline}          &   \multicolumn{2}{c}{NeuroSymbolic}         \\
    \cmidrule{3-11}
    Benchmark & &  Score & Steps    &  Score & Steps                 &   &       Score &  Steps   &   Score    &   Steps   \\
    \midrule
    MapReader   & &   0.02 & 50       &  0.02 & 50                      &  &       0.71        & 27        &   \textbf{1.00}        & \textbf{10}           \\
    Arithmetic  & &   0.17 & 10       &  0.14 & 7                       &  &       0.56        & 5         &   \textbf{1.00}        & \textbf{5}        \\
    Sorting     & &   0.03 & 21       &  0.03 & 18                      &  &       0.72        & 7         &   \textbf{0.98}        & \textbf{8}        \\
    TWC         & &   0.57 & 27       &  0.37 & 34                        &  &       0.90        & 6        &   \textbf{0.97}        & \textbf{3}        \\
    \midrule
    Average     & &   0.20 & 27       &  0.14 & 27                       &  &       0.72        & 11        &   \textbf{0.99}        & \textbf{7}        \\

    \bottomrule
    \end{tabular}
    \caption{Average model performance across 100 games in the unseen test set.  \textit{Scores} are normalized to between 0 and 1 (higher is better), while \textit{steps} represents the number of steps an agent takes in the environment (lower is better).  Neurosymbolic performance reflects when models have access to symbolic modules in their action space. }
    \label{tab:results}
\end{table*}

\section{Models}
In this section, we introduce the reinforcement learning and behavior cloning agents used in our experiments. Additional details and hyperparameters are provided in \textsc{Appendix~\ref{sec:exp}}.

{\flushleft\textbf{Deep Reinforcement Relevance Network (DRRN):}} The DRRN \cite{he-etal-2016-deep-reinforcement} is a fast and strong reinforcement learning baseline that is frequently used to deliver near state-of-the-art performance in a variety of text games \cite[e.g.][]{xu2020deep,yao-etal-2020-keep,wang2022scienceworld}. At each step, the DRRN separately encodes the observation and candidate actions using several GRUs \cite{cho-etal-2014-properties}. A Deep Q-Network is then used to estimate Q-values for each \textit{(observation, candidate action)} pair. The candidate action with the highest predicted Q-value will be chosen as the next action.


{\flushleft\textbf{Behavior Cloning:}} Behavior cloning \cite{ijcai2018p687} is a form of imitation learning similar to the Decision Transformer \cite{chen2021decision} that models reinforcement learning as a sequence-to-sequence problem, predicting the next action given a series of previous observations.  We follow the strategy of Ammanabrolu et al.~\shortcite{ammanabrolu-etal-2021-motivate} in adapting behavior cloning to text games, where the model input at step $t$ includes the task description, current state observation, previous action, and previous state observation $(d, o_t, a_{t-1}, o_{t-1})$. During training, the agent is fine-tuned on gold trajectories, where the training target is to generate action $a_t$ from the gold trajectories. During evaluation, the agent performs inference online in the text game environment.  For experiments reported here, we used a T5-base model \cite{JMLR:v21:20-074}.

\subsection{Oracle Agents and Gold Trajectories}
To generate training data for the behavioral cloning model, we implement oracle agents that generate optimal and generalizable solution trajectories for each benchmark.  For example, an oracle agent for an arithmetic game always reads the math problem, picks up the object with the same quantity as the math problem answer, then places that object in the answer box.  For experiments using symbolic modules, we further insert appropriate module actions when the agent requires that information to complete the next step -- for example, using the calculator module after reading the math problem in the arithmetic game.



\section{Results and Discussion}
The results of both DRRN and behavior cloning experiments across each benchmark are shown in Table~\ref{tab:results}. We report the average model performance across 100 games in the unseen test set.  The DRRN achieves a low average performance of 0.20 without modules, while adding symbolic modules into the action space does not improve performance.  In contrast, the behavior cloned T5 model has a moderate average performance of 0.72 without modules, while adding symbolic modules increases average task performance to 0.99, nearly solving each task. Symbolic modules also make the behavior cloned agent more efficient, reducing the average steps required to complete the tasks from 11 to 7, matching oracle agent efficiency.


{\flushleft\textbf{Why does behavior cloning perform well?}} 
The baseline behavior cloned transformer achieves moderate overall performance, likely owing at least in part due to its use of gold trajectories for training. Large pretrained transformers contain a variety of common sense knowledge and reasoning abilities \cite{zhou2020evaluating,liu2022generated} which likely contributes to the high performance on TWC, where the model only needs to match objects with their common sense locations.  In contrast, while transformers have some arithmetic abilities, their accuracy tends to vary with the frequency of specific tokens in the training data \cite{razeghi2022impact}, likely causing the modest performance on the Arithmetic game.  Here, we show that instead of increasing the size of training data, transformers can be augmented with symbolic modules that perform certain kinds of reasoning with high accuracy.  Compared to the DRRN, the presence of gold trajectories for training allows the behavior cloned transformer to efficiently learn how to capitalize on the knowledge available from those  modules.

{\flushleft\textbf{Why does the DRRN perform poorly?}} 
We hypothesize that two considerations make these tasks difficult for the Deep Reinforcement Relevance Network.
The model frequently tries to select actions that lead to immediate reward (such as immediately picking the correct number of objects in the arithmetic game), without having first done the prerequisite actions (like reading or solving the math problem) that would naturally lead it to select that action.  This creates an ungeneralizable training signal, causing the model to fail to learn the task.
In addition, the action spaces for each game are generally large -- baseline games contain between 5 and 30 possible valid actions at each step (see Table~\ref{tab:environment-statistics} in the \textsc{Appendix}), resulting in up to 24 million possible trajectories up to 5 steps, which is challenging to explore.  Inspired by \citet{Liu2022AskingFK}, our future work will aim to overcome these limitations, and allow reinforcement learning models to learn to efficiently and effectively exploit information from symbolic modules. 

{\flushleft\textbf{How does performance compare against other agents?}} 
While most environments used in this work are new, \textsc{TextWorld Common Sense} is an existing benchmark.  Figure~\ref{tab:twc-comparison} compares the Neurosymbolic Behavior Cloned Transformer against recent models that use a combination of reinforcement learning, logic, knowledge resources, and case-based reasoning.  While the performance is not directly comparable -- here, we use the \textsc{TextWorldExpress} reimplementation of TWC with supervised learning, while other models use the original implementation with a mix of reinforcement learning and case-based reasoning -- we can make the high-level observation that the performance of both the baseline and Neurosymbolic Behavior Cloned Transformer meets or exceeds the scores of previous models, while generating paths that are more efficient -- by a factor of up to 7x. 

%
%
\begin{table}[t]
    \centering
    \footnotesize
    \begin{tabular}{lcc}
    \toprule
    Model   &   Score   &   Steps   \\
    \midrule
    SceneIt \cite{murugesan2022eye}              &   0.88    &   20  \\
    Bike+CBR \cite{atzeni2022casebased}          &   0.93    &   17  \\
    SceneGraph \cite{tanaka2022commonsense}      &   0.91    &   17  \\
    IG \cite{basu2021hybrid}                     &   0.92    &   13  \\    
    \midrule
    BCT Baseline (Ours)                         &   0.90  &   6  \\
    \textbf{BCT+NeuroSymbic (Ours)}             &   \textbf{0.97}  &   \textbf{3}  \\
    \bottomrule
    \end{tabular}
    \caption{A comparison of performance on TWC on unseen games on the ``easy'' setting.  Note that models may not be directly comparable, as this work uses the \textsc{TextWorldExpress} reimplementation of TWC, and supervised learning.  \textit{Scores} are normalized to between 0 and 1 (higher is better), while \textit{steps} represents the number of steps an agent takes in the environment (lower is better). }
    \label{tab:twc-comparison}
\end{table}

\section{Conclusion}
In this paper, we present an approach to neurosymbolic reasoning for text games using action space injection that can be easily adapted to existing text game environments.  For models that are capable of exploiting the information provided by the symbolic modules, this technique allows agents to inexpensively augment their reasoning skills to solve more complex tasks.  We empirically demonstrate this approach can substantially increase task performance on four benchmark games using a behavior cloned transformer. 




\section*{Limitations}
Two assumptions highlight core limitations in the scope of our results for augmenting models with neurosymbolic reasoning: the privileged access to a list of valid actions, and the use of gold trajectories for training the behavior cloned transformer.

{\flushleft\textbf{Valid Actions:}} One of the central challenges with text games is that the space of possible action utterances is large, and text game parsers recognize only a subset of possible actions (e.g. \textit{take apple on the table}) while being unable to successfully interpret a broader range of more complex utterances (e.g. \textit{take the red fruit near the fridge}).  As a result, nearly all contemporary models \cite[e.g.][]{ammanabrolu2020graph, NEURIPS2020_1fc30b9d,murugesan2021textbased} make use of the \textit{valid action aid} \cite{Hausknecht-et-al-2022}, where at a given step the model is provided with an exhaustive list of possible valid actions from the environment simulator, from which one action is chosen.  The models presented here similarly use this aid. The DRRN functions essentially as a ranker to select the most probable next action.  The behavior cloned transformer generates a candidate action that is aligned using cosine similarity with the list of valid actions, where the action with the highest overlap is chosen as the next action.  Overcoming the \textit{valid action aid} will generally require either more complex simulation engines capable of interpreting a wider variety of intents from input actions, or models that learn sets of valid actions from a large amount of training data -- though these generally demonstrate lower performance than those using valid actions \cite[e.g.][]{yao-etal-2020-keep}. 

{\flushleft\textbf{Gold Trajectories:}} In this work we demonstrate a substantial improvement in the performance of a behavior cloned transformer when augmented with neurosymbolic reasoning, but this requires the use of gold trajectories demonstrating the use of those symbolic modules.  Gold training data is not available in many reinforcement learning applications, and the model comparison we perform (DRRN versus behavior cloning) is meant to highlight the capacity for the behavior cloned model to learn to make use of symbolic modules through gold demonstrations, rather than to suggest the DRRN is incapable of this.  In future work, we aim to develop training procedures to allow models that do not have the benefit of using gold trajectories to make use of symbolic modules. 

\section*{Ethics Statement}

{\flushleft\textbf{Broader Impacts:}} As noted by \citet{ammanabrolu2021learning}, the ability to perform long-term multi-step reasoning in complex, interactive, partially-observable environments has downstream applications beyond playing games.  Text games are platforms upon which to explore interactive, situated communication such as dialogue. Although reinforcement learning is applicable to many sequential decision making domains, our setting is most relevant to creating agents that affect change via language. This mitigates physical risks prevelant in robotics, but not cognitive and emotional risks, as any system capable of generating natural language is capable of biased language use~\cite{sheng-etal-2021-societal}.

{\flushleft\textbf{Intended Use:}} The method described in this paper involves fine-tuning a large pretrained transformer model.  The data generated for fine-tuning was generated by gold agents, and not collected from human participants.  The trained models are intended to operate on these benchmark tasks that assess reasoning capacities in navigation, arithmetic, and other common sense competencies.  Large language models have been shown to exhibit a variety of biases \cite[e.g.][]{nadeem-etal-2021-stereoset} that may cause unintended harms, particularly (in the context of this work) in unintended use cases. 

{\flushleft\textbf{Computation Time:}} Training large models can involve a large carbon footprint \cite{strubell-etal-2019-energy}, or decrease the availability of a method due to the barriers in accessing high performance compute resources.  The proposed technique can reduce the need for large models by augmenting smaller models with more complex reasoning through symbolic modules.  The behavior cloning experiments achieve strong performance with T5-base, highlighting the capacity of modest models that can be run with workstation GPUs to be better exploited for complex reasoning tasks.

\bibliography{anthology,custom}
\bibliographystyle{acl_natbib}

\appendix

%
%
\begin{table*}[t!]
    \centering
    \footnotesize
    \begin{tabular}{lp{5mm}cccp{5mm}ccc}
    \toprule
                & & \multicolumn{3}{c}{No Modules}    & &   \multicolumn{3}{c}{With Symbolic Modules}                    \\
    \cmidrule{3-9}
    Benchmark & &  Min & Avg    &  Max                &  &       Min & Avg  &   Max      \\
    \midrule
    MapReader   & &   4    &  6.2   &  22               &  &       6        &   9.3   &   26                  \\
    Arithmetic  & &   9    &  14.3  &  52               &  &       17       &  21.5   &  53            \\
    Sorting     & &   5    &  9.3   &  35               &  &       7        &  11.0   &   37            \\
    TWC         & &   3    &  6.3   &  14               &  &       544      &  547.8  &   561            \\
    \midrule
    Average     & &   5.3  &  9.0   &  30.8             &  &       143.5    &  147.4  &   169.3            \\

    \bottomrule
    \end{tabular}
    \caption{The minimum, mean, and maximum number of valid actions per step, for each benchmark.  Values represent averages determined using a random agent that is run to 50 steps for on 10 training episodes per benchmark. }
\label{tab:environment-statistics}
\end{table*}



\section{Appendix: Experiment Details}
\label{sec:exp}

\subsection{Training and evaluation sets} 
For each game, we randomly generate 100 parametric variations for each of the train, development, and test sets. To encourage and evaluate generality, problems are unique across sets -- for example, arithmetic problems (for the Arithmetic game) or task objects (for TWC) found in the training set are not found in the development or test sets.

\subsection{Hyperparameters}
Following standard practice (e.g. \cite{wang2022scienceworld, xu2020deep, Hausknecht-et-al-2022}), the DRRN models are trained for 100k steps. We parallelly train DRRN on 16 environment instances with five different random seeds and the average results are reported. The behavior cloned transformers are trained for between 2 and 20 epochs, with the best model (as evaluated on the development set) used for evaluating final performance on the test set.  
Trained models are evaluated on all 100 parametric variations in the development or test set. 
Environments are limited to 50 steps, such that if the agent exceeds this many steps without reaching an end state, the score at the last step is taken to be the final score, and the environment resets.  Model training time varied between 1 hour and 12 hours, with the TWC model that includes a large number of symbolic module actions requiring the largest training time.

\subsection{Implementation details}
We make use of an existing DRRN implementation\footnote{\url{https://github.com/microsoft/tdqn}} and adapted it to the \textsc{TextWorldExpress} environment. At each step, the current game state observation, task description, inventory information, and the current room description are concatenated into one string and encoded by a GRU. All candidate actions are encoded by another GRU. The Q-value of each encoded (observation, candidate action) pair is then estimated by a Q-network consists of two linear layers. During training, the next action is sampled from all candidate actions based on the estimated Q-values. During evaluation, the action with the highest estimated Q-value is chosen as the next action.

For the behavior cloned transformer, the input string of the T5 model at step $t$ are formatted as:\\


\noindent $d$ \token{</s>} \token{OBS} $o_t$ \token{</s>} \token{INV} $o^{inv}_t$ \token{</s>} \token{LOOK} $o^{look}_t$ \token{</s>} \token{<extra\_id\_0>} \token{</s>} \token{PACT} $a_{t-1}$ \token{</s>} \token{POBS} $o_{t-1}$ \token{</s>} ~\\

\noindent where $d$ is the task description, \token{</s>} and \token{<extra\_id\_0>} are the special tokens for separator and mask for text to generate used by the T5 model, \token{OBS}, \token{INV}, \token{LOOK}, \token{PACT}, and \token{POBS} are the special tokens representing observation $o_t$, inventory information $o^{inv}_t$, the current room description obtained by the ``look around'' action $o^{look}_t$, previous action $a_{t-1}$, and previous observation $o_{t-1}$, respectively. We use beam search to generate the top 16 strings from the T5 model, and choose the first string that is a valid action as the action to take.  In the case where the model does not generate an exact match, we use cosine similarity to pick the valid action that has the highest unigram overlap with an action generated by T5. 

\section{Environments and Symbolic Modules}
\label{sec:env}

{\flushleft\textbf{Action Space:}} The number of valid actions per step for each benchmark is shown in Table~\ref{tab:environment-statistics}, with these values collected by a random agent that runs for 50 steps across 10 training episodes. 
Environments contain an average of 9 valid actions per step (range 5 to 30), not including actions injected from the symbolic modules. After adding the module actions, the action space becomes up to twice as large for the MapReader, Arithmetic, and Sorting benchmarks.  The knowledge base module for TWC adds approximately 530 actions, substantially increasing the action space.

\subsection{Text World Common Sense (TWC)}

%
%
\begin{table}[t!]
\begin{center}
\scriptsize
\setlength{\tabcolsep}{3pt}
\begin{tabular}{p{0.98\linewidth}} 
\toprule
    \textbf{Text World Common Sense (TWC) Game} \\
\midrule
\textbf{Task Description:} Your task is to pick up objects, then place them in their usual locations in the environment.\\
\\
You are in the corridor. In one part of the room you see \underline{a shoe cabinet} that is closed. There is also \textbf{a white coat}. You also see \underline{a key holder}, that has nothing on it. In another part of the room you see \underline{a hat rack}, that has nothing on it. In one part of the room you see \textbf{a coat hanger}, that has nothing on it. There is also \underline{a umbrella stand}, that has nothing on it.\\
Inventory:\\ 
~~Your inventory is currently empty.\\
\textit{> \textcolor{cerulean}{query \textbf{white coat}}} \\
~\\
\textcolor{cerulean}{The results are:}\\
~~\textcolor{cerulean}{\textbf{white coat} located \textbf{coat hanger}}\\
~~\textcolor{cerulean}{\textbf{white coat} located wardrobe} \\
Inventory: \\
~~Your inventory is currently empty.\\
\textit{> take \textbf{white coat}} \\
~\\
You take the \textbf{white coat}.\\
Inventory: \\
~~\textbf{a white coat}\\
\textit{> put \textbf{white coat} in \textbf{coat hanger}} \\
~\\
\textit{Game completed.}\\

\bottomrule

\end{tabular}
\caption{An example of a Text World Common Sense (TWC) game. \textit{Actions} for the solution are italicized.  Relevant \textbf{task objects} and \textbf{target locations} are bolded, while \underline{distractor locations} are underlined. Symbolic module actions and their responses are highlighted in \textit{\textcolor{cerulean}{blue}}.
\label{tab:twc-example}}
\end{center}
\vspace{-6mm}
\end{table}

A benchmark common sense reasoning task where agents must collect objects from the environment (e.g. \textit{dirty socks}) and place them in their canonical locations (e.g. \textit{washing machine}).  We use the \textsc{TextWorldExpress} implementation of the original environment \cite{murugesan2021textbased}, in a setting analogous to easy difficulty, where the agent must pick and place one misplaced item in the environment.  The agent receives a score of 0.5 for picking up the object, and 1.0 for placing it in the correct location. Table~\ref{tab:twc-example} shows an example of the TWC Game in which the agent places a clean shirt in the wardrobe with the help of querying the common sense location of a clean shirt. 
 
\textit{Knowledge Base Module:} We pair TWC with a symbolic module that provides queries to a simple knowledge base of \textit{<object, hasCanonicalLocation, container>} triples.  The symbolic module generates and accepts actions of the form \textsc{query <query tokens>}, where \textsc{<query tokens>} corresponds to all object and container names in the knowledge base.  This results in an increase of approximately 530 actions in the action space.

\begin{table}[t!]
\begin{center}
\scriptsize
\setlength{\tabcolsep}{3pt}
\begin{tabular}{p{0.98\linewidth}} 
\toprule
    \textbf{Arithmetic Game} \\
\midrule
\textbf{Task Description:} Your first task is to solve the math problem. Then, pick up the item with the same quantity as the math problem answer, and place it in the box.\\
\\
You are in the kitchen. In one part of the room you see a fridge that is closed. There is also a dining chair that has \underline{11 tangerines}, \underline{33 papayas}, \underline{242 strawberries}, and \underline{20 peaches} on it. You also see a box, that is empty. In another part of the room you see a math problem. In one part of the room you see a dishwasher that is closed. There is also a trash can that is closed. You also see an oven. In another part of the room you see a cutlery drawer that is closed. In one part of the room you see a stove. There is also a kitchen cupboard that is closed. You also see a counter that has \underline{6 oranges}, and \textbf{2 bananas} on it.\\
Inventory:\\ 
~~Your inventory is currently empty.\\
\textit{> take math problem} \\
~\\
You take the math problem. \\
Inventory: \\
~~a math problem\\
\textit{> read math problem} \\
~\\
Your task is to solve the following math problem: divide 22 by 11. Then, pick up the item with the same quantity as the answer, and place it in the box.\\
Inventory: \\
~~a math problem\\
\textit{> \textcolor{cerulean}{div 22 11}} \\
~\\
\textcolor{cerulean}{The result of dividing 22 by 11 is 2.} \\
Inventory: \\
~~a math problem\\
\textit{> take \textbf{2 bananas}} \\
~\\
You take the \textbf{2 bananas}. \\
Inventory: \\
~~a math problem\\
~~\textbf{2 bananas}\\
\textit{> put \textbf{2 bananas} in box} \\
~\\
\textit{Game completed.}\\

\bottomrule

\end{tabular}
\caption{An example Arithmetic game. \textit{Actions} for the solution are italicized.  Relevant \textbf{task objects} are bolded, while \underline{distractor objects} are underlined. Symbolic module actions and their responses are highlighted in \textit{\textcolor{cerulean}{blue}}.
\label{tab:arithmetic-example}}
\end{center}
\vspace{-6mm}
\end{table}

\subsection{Arithmetic Game}
The Arithmetic game requires agents to read a math problem, solve it, then perform a pick-and-place task based on the answer.  For example, the agent may read the math problem (\textit{``Take the bundle of objects that is equal to 3 multiplied by 6, and place them in the box''}), and must then perform the arithmetic then \textit{take 18 apples} and place them in the answer box.  Distractor objects are populated corresponding to performing the arithmetic incorrectly (for example, including 3 oranges, corresponding to subtracting 3 from 6, and 2 pears, corresponding to 6 divided by 3), with the condition that results are positive integer values.  Agents receive a score of 0.5 for picking up the correct object, and 1.0 for completing the task successfully.
 An example playthrough of the Arithmetic game is in Table~\ref{tab:arithmetic-example}.
 
\textit{Arithmetic Module:} We pair the Arithmetic game with an Arithmetic module that adds actions for addition, subtraction, multiplication, and division.  To reduce the complexity of the action space, only actions with arguments from the current math problem are enumerated (e.g. \textit{add 3 6, sub 3 6, sub 6 3, mul 3 6, div 3 6, div 6 3}).

%
%
\begin{table}[t!]
\begin{center}
\scriptsize
\setlength{\tabcolsep}{3pt}
\begin{tabular}{p{0.98\linewidth}} 
\toprule
    \textbf{Sorting Game} \\
\midrule
\textbf{Task Description:} Your task is to sort objects by quantity.  First, place the object with the smallest quantity in the box.  Then, place the objects with the next smallest quantity in the box, and repeat until all objects have been placed in the box.\\
\\
You are in the kitchen. In one part of the room you see a fridge that is closed. There is also a counter that has \textbf{15kg of cedar}, and \textbf{21kg of marble} on it. You also see a dining chair that has \textbf{25g of oak}, and \textbf{47g of brick} on it. In another part of the room you see a box, that is empty. In one part of the room you see a dishwasher that is closed. There is also a trash can that is closed. You also see an oven. In another part of the room you see a cutlery drawer that is closed. In one part of the room you see a stove. There is also a kitchen cupboard that is closed.\\
Inventory:\\ 
~~Your inventory is currently empty.\\
\textit{> \textcolor{cerulean}{sort ascending}} \\
~\\
\textcolor{cerulean}{The observed items, sorted in order of increasing quantity, are: \textbf{25 g of oak}, \textbf{47 g of brick}, \textbf{15 kg of cedar}, \textbf{21 kg of marble}.}\\
Inventory: \\
~~Your inventory is currently empty.\\
\textit{> take \textbf{25g of oak}} \\
~\\
You take the \textbf{25g of oak}.\\
Inventory: \\
~~\textbf{25g of oak}\\
\textit{> put \textbf{25g of oak} in box} \\
~\\
You put the \textbf{25g of oak} in the box.\\
Inventory:\\ 
~~Your inventory is currently empty.\\
\textit{> take \textbf{47g of brick}} \\
~\\
You take the \textbf{47g of brick}.\\
Inventory: \\
~~\textbf{47g of brick}\\
\textit{> put \textbf{47g of brick} in box} \\
~\\
You put the \textbf{47g of brick} in the box.\\
Inventory:\\ 
~~Your inventory is currently empty.\\
\textit{> take \textbf{15kg of cedar}} \\
~\\
You take the \textbf{15kg of cedar}.\\
Inventory: \\
~~\textbf{15kg of cedar}\\
\textit{> put \textbf{15kg of cedar} in box} \\
~\\
You put the \textbf{15kg of cedar} in the box.\\
Inventory:\\ 
~~Your inventory is currently empty.\\
\textit{> take \textbf{21kg of marble}} \\
~\\
You take the \textbf{21kg of marble}.\\
Inventory: \\
~~\textbf{21kg of marble}\\
\textit{> put \textbf{21kg of marble} in box} \\
~\\
\textit{Game completed.}\\

\bottomrule

\end{tabular}
\caption{An example Sorting game. \textit{Actions} for the solution are italicized. Relevant \textbf{task objects} are bolded. Symbolic module actions and their responses are highlighted in \textit{\textcolor{cerulean}{blue}}.
\label{tab:sorting-example}}
\end{center}
\vspace{-6mm}
\end{table}

\subsection{Sorting Game}
The sorting game is a pick-and-place game that presents an agent with 3 to 5 objects, and asks the agent to place them in an answer box one at a time based on order of increasing quantity.  To add complexity to the game, quantities optionally include units (e.g. \textit{5kg of copper, 8mg of steel, 2g of iron}) across measures of volume, mass, or length.  The agent score is the normalized proportion of objects sorted in the correct order, where perfect sorts receive a score of 1.0, and errors cause the score to revert to zero and the game to end. An example playthrough of the Sorting game is in Table~\ref{tab:sorting-example}.

\textit{Sorting Module:} The sorting module monitors observations for mentions of objects (nouns) that include quantities, while also interpreting and normalizing quantities based on known units.  The module injects two actions: sort ascending, and sort descending, that provides the user with a sorted list of objects.

%
%
\begin{table*}[t!]
\begin{center}
\scriptsize
\setlength{\tabcolsep}{3pt}
\begin{tabular}{p{0.45\linewidth}p{0.05\linewidth}p{0.45\linewidth}} 
\toprule
    \textbf{MapReader Game} \\
\midrule
\textbf{Task Description:} Your task is to take the coin that is located in \textbf{the laundry room}, and put it into the box found in \textbf{the foyer}. A map is provided, that you may find helpful.	&	&	\textit{continued...} \\
	&	&	\\
You are in \textbf{the foyer}. In one part of the room you see a box, that is empty.	&	&	You are in \textbf{the laundry room}. In one part of the room you see a coin. There \\
To the East you see the corridor.	&	&	is also a bench, that has nothing on it. You also see a washing machine \\
Inventory:	&	&	that is closed. In another part of the room you see a work table, that has \\
~~a map	&	&	nothing on it. In one part of the room you see a laundry basket, that has \\
> \textit{read map}	&	&	nothing on it. There is also a clothes drier that is closed. \\
~	&	&	To the South you see the corridor. \\
The map reads:	&	&	Inventory: \\
~~The living room connects to the backyard and corridor.	&	&	~~a map\\
~~The garage connects to the driveway.	&	&	> \textit{take coin}\\
~~\textbf{The laundry room} connects to the corridor.	&	&	~\\
~~The backyard connects to the living room, alley, kitchen and sideyard.	&	&	You take the coin.\\
~~The bedroom connects to the corridor.	&	&	Inventory: \\
~~The sideyard connects to the backyard and driveway.	&	&	~~a map\\
~~The kitchen connects to the bathroom, pantry and backyard.	&	&	~~a coin\\
~~The supermarket connects to the alley.	&	&	> \textit{\textcolor{cerulean}{next step to \textbf{foyer}}}\\
~~\textbf{The foyer} connects to the corridor.	&	&	~\\
~~The pantry connects to the kitchen.	&	&	\textcolor{cerulean}{The next location to go to is: corridor}\\
~~The driveway connects to the sideyard, alley and garage.	&	&	Inventory: \\
~~The street connects to the alley.	&	&	~~a map\\
~~The alley connects to the driveway, supermarket, street and backyard.	&	&	~~a coin\\
~~The bathroom connects to the kitchen.	&	&	> \textit{move south}\\
~~The corridor connects to the living room, foyer, bedroom and laundry	&	&	~\\
~~~~room.	&	&	You are in the corridor. In one part of the room you see a shoe cabinet that \\
Inventory:	&	&	is closed. There is also a key holder, that has nothing on it. You also see a \\
~~a map	&	&	hat rack, that has nothing on it. In another part of the room you see a coat \\
> \textit{\textcolor{cerulean}{next step to laundry room}}	&	&	hanger, that has nothing on it. In one part of the room you see a umbrella \\
~	&	&	stand, that has nothing on it. \\
\textcolor{cerulean}{The next location to go to is: corridor}	&	&	To the North you see \textbf{the laundry room}. To the South you see the living \\
Inventory: 	&	&	room. To the East you see the bedroom. To the West you see \textbf{the foyer}. \\
~~a map	&	&	Inventory: \\
> \textit{move east}	&	&	~~a map\\
~	&	&	~~a coin\\
You are in the corridor. In one part of the room you see a shoe cabinet that	&	&	~\\
is closed. There is also a key holder, that has nothing on it. You also see a	&	&	> \textit{\textcolor{cerulean}{next step to \textbf{foyer}}}\\
hat rack, that has nothing on it. In another part of the room you see a coat	&	&	\textcolor{cerulean}{The next location to go to is: \textbf{foyer}}\\
hanger, that has nothing on it. In one part of the room you see a umbrella	&	&	Inventory: \\
stand, that has nothing on it.	&	&	~~a map\\
To the North you see the laundry room. To the South you see the living	&	&	~~a coin\\
room. To the East you see the bedroom. To the West you see \textbf{the foyer}.	&	&	> \textit{move west}\\
Inventory: 	&	&	~\\
~~a map	&	&	You are in \textbf{the foyer}. In one part of the room you see a box, that is empty. \\
> \textit{\textcolor{cerulean}{next step to \textbf{laundry room}}}	&	&	To the East you see the corridor. \\
~	&	&	Inventory: \\
\textcolor{cerulean}{The next location to go to is: \textbf{laundry room}}	&	&	~~a map\\
Inventory: 	&	&	~~a coin\\
~~a map	&	&	> \textit{put coin in box}\\
> \textit{move north}	&	&	~\\
~	&	&	\textit{Game completed.}\\
\bottomrule

\end{tabular}
\caption{An example of a MapReader game. \textit{Actions} for the solution are italicized.  The \textbf{starting location} and the \textbf{target location} are bolded. Symbolic module actions and their responses are highlighted in \textit{\textcolor{cerulean}{blue}}. 
\label{tab:mapreader-example}}
\end{center}
\vspace{-6mm}
\end{table*}

\subsection{MapReader Game}
MapReader is a navigation oriented pick-and-place game similar to Coin Collector \cite{Yuan2018CountingTE}, with the added complexity that the agent is provided with a map of the environment that can be used to more efficiently navigate.  Environments and their maps are randomly generated to contain up to 15 locations drawn from 50 locations in Coin Collector.  The agent begins in a randomly chosen location, and is asked to move to a target location (e.g. \textit{the kitchen}), take a coin, then return to the starting location and place it in a box.  Target locations are randomly chosen to be between 1 and 4 steps from the starting location.  The most efficient solution method is to read the map, determine the shortest path between the agents current location and target location, follow that path to retrieve the coin, then follow the path in reverse to return the coin to the starting location.  The agent receives a score of 0.5 for retrieving the coin, and 1.0 for placing the coin in the box at the start location.
An example of the MapReader game is shown in Table~\ref{tab:mapreader-example}.

\textit{Navigation Module:} We pair MapReader with a navigation module that scrapes the environment for both complete map information (obtained if the agent chooses to read the map), as well as partial information such as the current location (e.g. \textit{``You are in the kitchen''}) and connecting locations (e.g. \textit{``To the north you see the living room''}) that can be used to incrementally build a map. The module adds actions that, if selected, provide the next step in the shortest path to navigate to all known locations in the environment (e.g. \textit{next step to living room}, \textit{next step to garage}, ...). 


\end{document}